\pdfoutput=1
\documentclass[10pt,twocolumn,letterpaper]{article}

\usepackage{cvpr}
\usepackage{times}
\usepackage{epsfig}
\usepackage{graphicx}
\usepackage{amsmath}
\usepackage{amssymb}

\usepackage[table]{xcolor}
\usepackage{color}
\usepackage{colortbl}
\usepackage{caption}
\usepackage{subcaption}
\usepackage{tabularx}
\usepackage{array}
\usepackage{multirow}
\usepackage{float}

\usepackage[breaklinks=true,bookmarks=false]{hyperref}

\cvprfinalcopy 

\def\cvprPaperID{****} 

\begin{document}
\newcommand*\samethanks[1][\value{footnote}]{\footnotemark[#1]}

\title{Convolutional Gated Recurrent Networks for  Video Segmentation}

\author{Mennatullah Siam \thanks{Authors contributed equally}
\qquad Sepehr Valipour \samethanks
\qquad Martin Jagersand
\qquad Nilanjan Ray\\
University of Alberta\\
{\tt\small \{mennatul,valipour,mj7,nray1\}@ualberta.ca}
}

\maketitle


\begin{abstract}
Semantic segmentation has recently witnessed  major progress, where fully convolutional neural networks have shown to perform well. However, most of the previous work focused on improving single image segmentation. To our knowledge, no prior work has made use of temporal video information in a recurrent network. In this paper, we introduce a novel approach to implicitly utilize temporal data in videos for online semantic segmentation. The method relies on a fully convolutional network that is embedded into a gated recurrent architecture. This design receives a sequence of consecutive video frames and outputs the segmentation of the last frame. Convolutional gated recurrent networks are used for the recurrent part to preserve spatial connectivities in the image. Our proposed method can be applied in both online and batch segmentation. This architecture is tested for both binary and semantic video segmentation tasks. Experiments are conducted on the recent benchmarks in SegTrack V2, Davis, CityScapes, and Synthia.  Using recurrent fully convolutional networks improved the baseline network performance in all of our experiments. Namely, 5\% and 3\% improvement of F-measure in SegTrack2 and Davis respectively, 5.7\% improvement in mean IoU in Synthia and 3.5\% improvement in categorical mean IoU in CityScapes. The performance of the RFCN network depends on its baseline fully convolutional network. Thus RFCN architecture can be seen as a method to improve its baseline segmentation network by exploiting spatiotemporal information in videos.
\end{abstract}

\section{Introduction}
\textbf{Semantic segmentation}, which provides pixel-wise labels, has witnessed a tremendous progress recently. As shown in \cite{long2015fully}\cite{noh2015learning}\cite{zheng2015conditional}\cite{visin2015reseg}, it outputs dense predictions and partitions the image to semantically meaningful parts. It has numerous applications including autonomous driving\cite{zhang2013understanding}\cite{ros2016synthia}\cite{cordts2016cityscapes}, augmented reality\cite{miksik2015semantic} and robotics\cite{vineet2015icra} \cite{wolf2016enhancing}. The work in \cite{long2015fully} presented a fully convolutional network and provides a method for end-to-end training of semantic segmentation. It yields a coarse heat-map followed by in-network upsampling to get dense predictions. Following the work of fully convolutional networks, many attempts were made to improve single image semantic segmentation. In \cite{noh2015learning} a full deconvolution network is presented with stacked deconvolution layers. The work in \cite{visin2015reseg} provided a method to incorporate contextual information using recurrent neural networks. However, one missing element is that the real-world is not a set of still images. In real-time camera or recorded video, much information is perceived from temporal cues. For example, the difference between a walking or standing person is hardly recognizable in still images but it is obvious in a video.
\begin{figure}[ht!]
     \centering
    \includegraphics[width=0.5\textwidth]{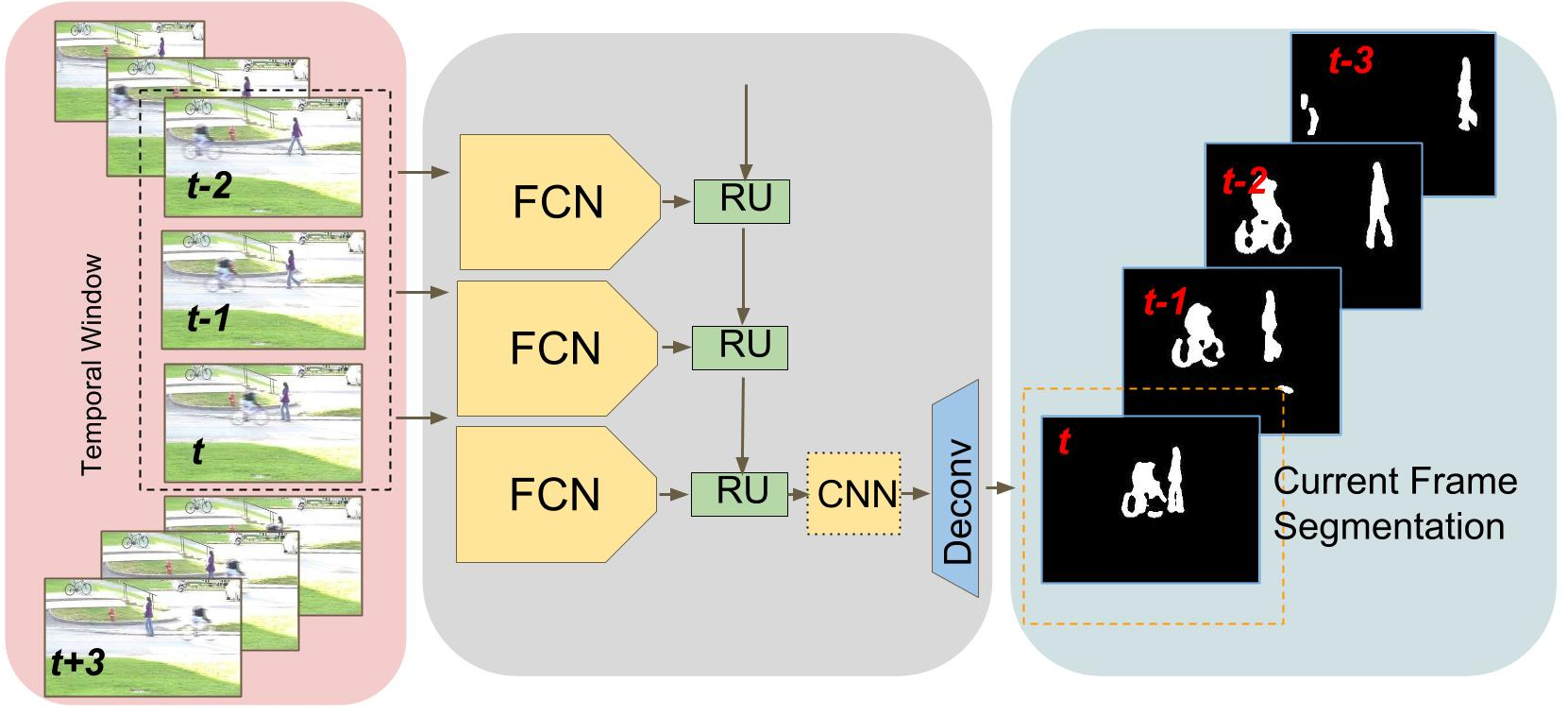}
    \caption{Overview of the Proposed Method of Recurrent FCN. The recurrent part is unrolled for better visualisation}
    \label{fig:overprop}
\end{figure}

\textbf{Video segmentation} has been extensively investigated using classical approaches. The work in \cite{perazzibenchmark} reviews the literature in binary video segmentation. It mainly focuses on semi-supervised approaches\cite{badrinarayanan2010label} \cite{perazzi2015fully}\cite{ramakanth2014seamseg} that propagate the labels in one or more annotated frames to the entire video. In \cite{pavel2015recurrent} a method that uses a combination of Recurrent Neural Networks (RNN) and CNN for RGB-D video segmentation is presented. However, their proposed architecture is difficult to train because of the vanishing gradient. It does not utilize pre-trained networks and it cannot process large images as number of their parameters is quadratic with respect to the input size. 

\textbf{Gated Recurrent Architectures} alleviate the vanishing or exploding gradients problem in recurrent networks. Long Short Term Memory (LSTM) architecture is one of the earliest attempt to design these networks\cite{hochreiter1997long}. They are successfully employed in many tasks. Captioning and text processing in particular \cite{johnson2015densecap} \cite{donahue2015long}. Gated Recurrent Unit (GRU) \cite{cho2014properties} is a more recent gated architecture. It is shown that GRU has similar performance to LSTM but with reduced number of gates thus fewer parameters\cite{chung2014empirical}. The main bottleneck with these previous architectures is that they only work with vectors and therefore, do not preserve spatial information in images or feature maps. In\cite{ballas2015delving} convolutional GRU is introduced for learning spatio-temporal features from videos and used for video captioning and action recognition.

Inspired by these methods we design a gated recurrent FCN architecture to solve many of the shortcomings of the previous approaches. Contributions include: 
\begin{itemize}
\item A novel architecture that can incorporate temporal data into FCN for video segmentation. A Convolutional Gated Recurrent FCN architecture is designed to efficiently utilize spatiotemporal information.
\item An end-to-end training method for online video segmentation. 
\item An experimental analysis on video binary segmentation and video semantic segmentation is presented on recent benchmarks. 
\end{itemize}
An overview of the suggested method is shown in Figure \ref{fig:overprop}. Where a sliding window of input video frames is fed to the recurrent fully convolutional network(RFCN) and resulted in the segmentation of the last frame. The paper is structured as follows. In Section \ref{background} necessary background is discussed. The proposed method is presented in details in section \ref{method}. Section \ref{experiments} presents experimental results and discussion on recent benchmarks. Finally, section \ref{conclusion} summarizes the article and presents potential future directions.


\section{Background} \label{background}
This section will review FCN and RNN which will be repeatedly referred to throughout the article.

\subsection{Fully Convolutional Networks (FCN)}
Convolutional neural networks that are initially designed with image classification tasks in mind. Later, it became apparent that CNN can also be used for segmentation by doing pixel-wise classification. However, dense pixel-wise labeling is extremely inefficient using regular CNN. In \cite{lin2015efficient} the idea of using a fully convolutional neural network that is trained for pixel-wise semantic segmentation is presented. In this approach, all the fully connected layers of CNN networks are replaced with convolutional layers. This design allows the network to accommodate any input size since it is not restricted to a fixed output size, fully connected layers. More importantly, now it is possible to get a course segmentation output (called heat-map) by only one forward pass of the network. 

This coarse map needs to be up-sampled to the original size of the input image. Simple bi-linear interpolation can be used however, an adaptive up-sampling is shown to have a better result. In \cite{lin2015efficient} a new layer with learnable filters that applies up-sampling within the network is presented. It is an efficient way to learn the up-sampling weights through back-propagation. These types of layers are commonly known as deconvolution. The filters of deconvolution layers can be seen as a basis to reconstruct the input image or just to increase the spatial size of feature maps. Skip architecture can be used for an even finer segmentation. In this architecture heat maps from earlier pooling layers are merged with the final heatmap for an improved segmentation. This architecture is termed as FCN-16s or FCN-8s based on the pooling layers that are used.

\subsection{Recurrent Neural Networks}

Recurrent Neural Networks\cite{vinyals2012revisiting} can be applied on a sequence if inputs and are able to capture the temporal relation between them. A hidden unit in each recurrent cell allows it to have a dynamic memory that is changing according to what it had hold before and the new input. The simplest recurrent unit can be modeled as in equation \ref{eq:rnn}. 
\begin{subequations}
\label{eq:rnn}
\begin{align}
    h_t &= \theta\phi(h_{t-1}) + \theta_x x_t \\
    y_t &= \theta_y \phi(h_t)
\end{align}    
\end{subequations}
Where, $h$ is the hidden unit, $x$ is the input, $y$ is the output, $t$ is the current time step and $\phi$ is the activation function.
 
When propagating the error in recurrent units, due to the chain law, we see that the derivative of each node is dependent on all of earlier nodes. This chain dependency can be arbitrary long based on length of the input vector. It was observed that it will cause vanishing gradient problem, especially for longer input vectors\cite{bengio1994learning}. Gated recurrent architectures have been proposed as a solution and they were empirically successful in many tasks. Two popular choices of these architectures are presented in this section. 

\subsubsection{Long Short Term Memory (LSTM)}

LSTM\cite{hochreiter1997long} utilizes three gates to control the flow of signal within the cell. These gates are input, output and forget gate and each of them has its own set of weights. These weights can be learned with back-propagation. At the inference stage, the values in the hidden unit changes based on the sequence of inputs that is has seen and can be roughly interpreted as a memory. This memory can be used for the prediction of the current state. Equations \ref{eq:lstm} shows how the gates values and hidden states are computed. $i_t$, $f_t$ and $o_t$ are the gates and $c_t$ and $h_t$ are the internal and the hidden state respectively.
\begin{subequations}
\begin{align}
    i_t &= \sigma(W_{xi}x_t + W_{hi}h_{t-1} + b_i) \\
    f_t &= \sigma(W_{xf}x_t + W_{hf}h_{t-1} + b_f) \\ 
    o_t &= \sigma(W_{xo}x_t + W_{ho}h_{t-1} + b_o) \\
    g_t &= \sigma(W_{xc}x_t + W_{hc}h_{t-1} + b_c) \\ 
    c_t &= f_t\odot c_{t-1}+i_t \odot g_t \\ 
    h_t &= o_t \odot \phi(c_t)
\end{align}
\label{eq:lstm}
\end{subequations}
\subsubsection{Gated Recurrent Unit (GRU)}
GRU uses the same gated principal of LSTM but with a simpler architecture. Therefore, it is not as computationally expensive as LSTM and uses less memory. Equations \ref{eq:gru} describe the mathematical model of the GRU. Here, $r_t$ and $z_t$ are the gates and $h_t$ is the hidden state.
\begin{subequations}
\label{eq:gru}
\begin{align}
    z_t &= \sigma(W_{hz} h_{t-1} + W_{xz} x_t + b_z)\\
    r_t &= \sigma(W_{hr} h_{t-1} + W_{xr} x_t + b_r)\\
    \hat{h_t} &= \Phi(W_h (r_t \odot h_{t-1}) + W_x x_t + b)\\
    h_t &= (1-z_t)\odot h_{t-1} + z \odot \hat{h_t} 
\end{align}
\end{subequations}

GRU is simpler than LSTM since the output gate is removed from the cell and the output flow is controlled by two other gates indirectly. The cell memory is also updated differently in GRU. LSTM updates its hidden state by summation over flow after input gate and forget gate. In the other hand, GRU assumes a correlation between memorizing and forgetting and controls both by one gate only $z_t$. 

\section{Method} \label{method}
An overview of the method is presented in Figure \ref{fig:overprop}. A recurrent fully convolutional network (RFCN) is designed that utilizes the spatiotemporal information for video segmentation. The recurrent unit in the network can either be LSTM, GRU or Conv-GRU (which is explained in \ref{conv_gru}). A sliding window over the video frames is used as input to the network. This allows on-line video segmentation as opposed to off-line batch processing. The window data is forwarded through the RFCN to yield a segmentation for the last frame in the sliding window. Note that the recurrent unit can be applied on the coarse segmentation (heat map) or intermediate feature maps. The network is trained in an end-to-end fashion using pixel-wise classification logarithmic loss. 
Two main approaches are explored in our method: (1) conventional recurrent units, and (2) convolutional recurrent units \ref{table:networks}. Specifically, four different network architectures under these two approaches are used as detailed in the following sections.

\subsection{Conventional Recurrent Architecture for Segmentation} 
\textbf{RFC-Lenet} is a fully convolutional version of Lenet. Lenet is a well known shallow network. Because it is common, we used it for baseline comparisons on synthetic data. We embed this model in a recurrent node to capture temporal data. The final network is named as RFC-Lenet in Table \ref{table:networks}.

The output of deconvolution a 2D map of dense predictions that is then flattened into 1D vector as the input to a conventional recurrent unit. The recurrent unit takes this vector for each frame in the sliding window and outputs the segmentation of the last frame (Figure \ref{fig:overprop}).
\newcolumntype{C}[1]{>{\centering\arraybackslash\hspace{0pt}}p{#1}}
\definecolor{lightgray}{gray}{0.5}
\begin{table*}[ht!]
\centering
\caption{Proposed networks in detail. $F(n)$ is a filter with size of $n\times n$. $P(n)$ indicates $n$ zero padding. $S(n)$ shows the stride in the convolution layer. $D(n)$ is number of feature maps generated by the layer $n$ (It is same as the previous layer if it is not mentioned).}
\label{table:networks}
\begin{tabular}{|c|C{4.5cm}||c|C{4.5cm}||c|C{5cm}|}
\hline
 \multicolumn{6}{|c|}{Network Architectures}  \\ \hline \hline
 \multicolumn{2}{|c||}{RFC-Lenet} &\multicolumn{2}{|c||}{ RFC-12s}  & \multicolumn{2}{|c|}{RFC-VGG} \\ \hline
 \multicolumn{2}{|c||}{input: 28$\times$28 }& \multicolumn{2}{|c||}{input: 120$\times$180} &  \multicolumn{2}{|c|}{input: 240$\times$360}\\ \hline
 \multirow{17}{*}{\rotatebox[origin=c]{90}{Recurrent Node} } & Conv: F(5), P(10), D(20)& \multirow{16}{*}{\rotatebox[origin=c]{90}{Recurrent Node} }& Conv: F(5), S(3), P(10), D(20) & \multirow{15}{*}{\rotatebox[origin=c]{90}{Recurrent Node} }& Conv: F(11), S(4), P(40), D(64) \\ \cline{2-2} \cline{4-4} \cline{6-6}
 & Relu& & Relu  & & Relu  \\ \cline{2-2} \cline{4-4} \cline{6-6}
 & Pool 2$\times$2 &  & Pool 2$\times$2 & & Pool 3$\times$3  \\ \cline{2-2} \cline{4-4} \cline{6-6}
 & Conv: F(5), D(50)& &   Conv: F(5), D(50)&  & Conv: F(5), P(2) D(256)\\ \cline{2-2} \cline{4-4} \cline{6-6}
 & Relu& & Relu  & & Relu  \\ \cline{2-2} \cline{4-4} \cline{6-6}
 & Pool(2$\times$2)& &  Pool(2$\times$2)& & Pool(3$\times$3) \\ \cline{2-2} \cline{4-4} \cline{6-6}
 & Conv: F(3), D(500)& &Conv: F(3), D(500)  & & Conv: F(3), P(1) D(256)   \\ \cline{2-2} \cline{4-4} \cline{6-6}
 & Relu& & Relu  & & Relu  \\ \cline{2-2} \cline{4-4} \cline{6-6}
 & Conv: F(1), D(1)& & Conv: F(1), D(1) & & Conv: F(3), P(1) D(256) \\ \cline{2-2} \cline{4-4} \cline{6-6}
 & \multirow{ 5}{*}{-} & & \multirow{ 5}{*}{-} & & Relu  \\ \cline{6-6}
 & & & & &  Conv: F(3), P(1) D(256)\\ \cline{6-6}
 & & & & & Relu\\ \cline{6-6}
 & & & & & Conv: F(3), D(512) \\ \cline{6-6}
 & & & & & Conv: F(3), D(128) \\  \cline{2-2} \cline{4-4} \cline{6-6}
 & DeConv: F(10), S(4)& & Flatten & & ConvGRU: F(3), D(128) \\ \cline{2-2} \cline{4-4} \cline{5-6}
 & Flatten& & GRU: W(100$\times$100) &\cellcolor{lightgray} & Conv: F(1), D(1)\\ \cline{2-2} \cline{3-4} \cline{6-6}
 & GRU: W(784$\times$784) & \cellcolor{lightgray} &  DeConv: F(10), S(4) & \cellcolor{lightgray}& DeConv: F(20), S(8)\\ \hline
 
\end{tabular}
\end{table*}

\textbf{RFC-12s} is another architecture that is used for baseline comparisons, to compare end-to-end and decoupled training as detailed in section \ref{experiments}. The RFC-Lenet architecture requires a large weight matrix in the recurrent unit since it processes vectors of the flattened full sized image. One way to overcome this problem is to apply the recurrent layer on the down-sampled heatmap before deconvolution. This leads to this second architecture termed as RFC-12s in Table \ref{table:networks}.
In this network, vectorized coarse output maps are given to the recurrent unit. The recurrent unit operates on a sequence of these coarse maps and produces a coarse map corresponding the last frame in the sequence. Later, the deconvolution layer generates dense prediction from the output the recurrent unit. In this way, the recurrent unit is allowed to work on much smaller vectors and therefore reduces the variance in the network.

\subsection{Convolutional Gated Recurrent Architecture (Conv-GRU) for Segmentation} \label{conv_gru}
\begin{figure*}[ht]
     \centering
    \includegraphics[width=\textwidth]{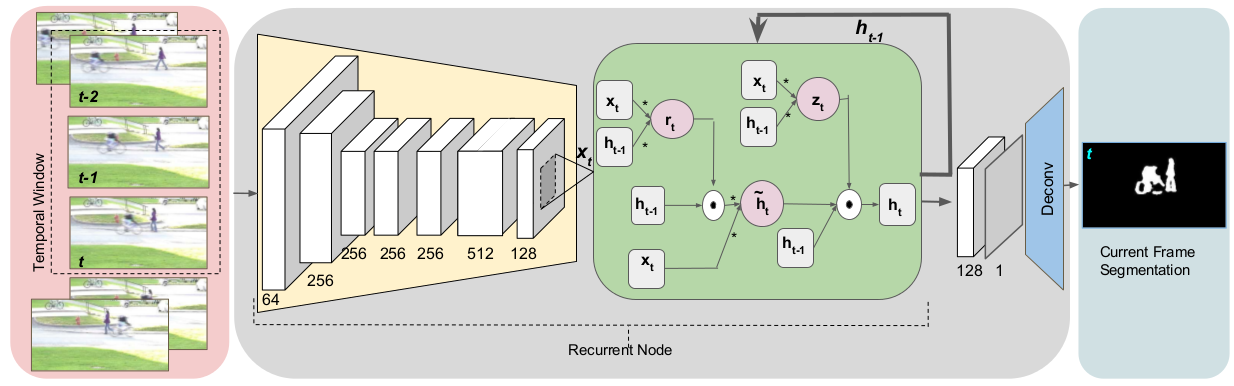}
    \caption{RFC-VGG architecture. A sequence of images is given as input to the network. The output of the embedded FC inside the recurrent unit is given to a Conv-GRU layer. One last convolutional layer maps the output of the recurrent unit into a coarse segmentation map. Then, the deconvolutional layer up-samples the coarse segmentation into a dense segmentation.}
    \label{fig:rfcvgg_detailed}
\end{figure*}

Conventional recurrent units are designed for processing text data not images. Therefore, using them on images without any modification causes two main issues. 1) The size of weight parameters becomes very large since vectorized images are large 2) Spatial connectivity between pixels are ignored.
For example, using a recurrent unit on a feature map with the spatial size of $h \times w$ and number of channels $c$ requires $c\times(h.w)^2$ number of weights. This will cause a memory bottleneck and inefficient computations. It will also create a larger search space for the optimizer, thus it will be harder to train.

Convolutional recurrent units, akin to regular convolutional layer, convolve three dimensional weights with their input. Therefore, to convert a gated architecture to a convolutional one, dot products should be replaced with convolutions. Equations \ref{eq:conv_gru} show this modification for the GRU. The weights are of size of $k_h \times k_w \times c \times f$ where $k_h$, $k_w$, $c$ and $f$ are kernel's height and width, number of input channels, and number of filters, respectively. Learning filters that convolve with the entire image instead of learning individual weights for each pixel, makes it much more efficient. This layer can be applied on either final heat map or intermediate feature maps.

\begin{subequations}
\label{eq:conv_gru}
\begin{align}
    z_t &= \sigma(W_{hz}\ast h_{t-1} + W_{xz}\ast x_t + b_z)\\
    r_t &= \sigma(W_{hr} \ast h_{t-1} + W_{xr} \ast x_t + b_r)\\
    \hat{h_t} &= \Phi(W_h \ast (r_t \odot h_{t-1}) + W_x \ast x_t + b)\\
    h_t &= (1-z_t)\odot h_{t-1} + z \odot \hat{h_t} 
\end{align}
\end{subequations} 
\textbf{RFC-VGG} in Table \ref{table:networks} is an example of this approach, where intermediate feature maps are fed into a convolutional gated recurrent unit. Then a convolutional layer converts its output to a heat map. It is based on VGG-F \cite{simonyan2014very} network. The reason for switching to the RFC-VGG architecture is to use pre-trained weights from VGG-F. Initializing weights of our filters by VGG-F trained weights, alleviates over-fitting problems as these weights are the result of extensive training on Imagenet dataset. The last two pooling layers are dropped from VGG-F to allow a finer segmentation with a reduced network. Figure \ref{fig:rfcvgg_detailed} shows the detailed architecture of RFC-VGG.

\textbf{RFCN-8s} is the recurrent version of FCN-8s architecture and is used in our semantic segmentation experiments. FCN-8s network is commonly used in many state of the art segmentation methods as it provides more detailed segmentation. It is loaded with a pre-trained with VGG-16 and it employs the skip architecture that combines pool3 and pool4 layers, with the final layer to have a finer segmentation. In RFCN-8s the convolutional gated recurrent unit is placed before pool3 layer where the skip connections start branching. 

\section{Experiments} \label{experiments}
This section describes the experimental analysis and results. First, the datasets are presented followed by the training method and hyper-parameters used. Then both quantitative and qualitative analyses are presented. All experiments are performed on our implemented open source library that supports convolutional gated recurrent architectures. The implementation is based on Theano \cite{Bastien-Theano-2012} and supports using different FCN architectures as a recurrent node. The key features of this implementation are: \textbf{(1)} The ability to use any arbitrary CNN or FCN architecture as a recurrent node. In order to utilize temporal information.
\textbf{(2)} Support for three gated recurrent architectures which are, LSTM, GRU, and Conv-GRU. \textbf{(3)} It includes deconvolution layer for in the network upsampling and supports skip architecture for finer segmentation. A public version of the code for the library along with the trained models will be published after the anonymous review.

\subsection{Datasets}
In this paper six datasets are used: 1) Moving MNIST. 2) Change detection\cite{goyette2012changedetection}. 3) Segtrack version 2\cite{li2013video}. 4) Densely Annotated VIdeo Segmentation (Davis) \cite{perazzibenchmark}. 5) Synthia\cite{ros2016synthia}. 6) CityScapes\cite{cordts2016cityscapes}

\textbf{Moving MNIST} dataset is synthesized from original MNIST by moving the characters in random but consistent directions. The labels for segmentation is generated by thresholding input images after translation. We consider each translated image as a new frame. Therefore we can have arbitrary length image sequences. 

\textbf{Change Detection Dataset\cite{goyette2012changedetection}} This dataset provides realistic, diverse set of videos with pixel-wise labeling of moving objects. The dataset includes both indoor and outdoor scenes. It focuses on moving object segmentation. In the foreground detection, videos with similar objects were selected such as humans or cars. Accordingly, we chose six videos: Pedestrians, PETS2006, Badminton, CopyMachine, Office, and Sofa. 

\textbf{SegTrack V2\cite{li2013video}} is a collection of fourteen video sequences with objects of interest manually segmented. The dataset has sequences with both single or multiple objects. In the latter case, we consider all the segmented objects as one and we perform foreground segmentation.

\textbf{Davis\cite{perazzibenchmark}} dataset includes fifty densely annotated videos with pixel accurate groundtruth for the most salient object. Multiple challenges are included in the dataset such as occlusions, illumination changes, fast motion, motion blur and nonlinear deformation. 

\textbf{Synthia\cite{ros2016synthia}} is a synthetic semantic segmentation dataset for urban scenes. It contains pixel level annotations for thirteen classes. It has over 200,000 images with different weather conditions (rainy, sunset, winter) and seasons (summer, fall). Since the dataset is large only a portion of it from Highway sequence for summer condition is used for our experiments.  

\textbf{CityScapes\cite{cordts2016cityscapes}} is a real dataset focused on urban scenes gathered by capturing videos while driving in different cities. It contains 5000 finely annotated 20000 coarsely annotated images for thirty classes. The coarse annotation includes segmentation for all frames in the video and each twentieth image in the video sequence is finely annotated. It provides various locations (fifty cities) and weather conditions throughout different seasons.

\subsection{Results}
The main experiments' setup includes using Adadelta  \cite{zeiler2012adadelta} for optimization as it gave much faster convergence than stochastic gradient descent. The loss function used throughout the experiments is the logistic loss, and the maximum number of epochs used for training is 500. The evaluation metrics used for the binary video segmentation is precision, recall, F-measure and IoU. Metrics formulation is shown in \ref{eq-prec}, \ref{eq-fmes} and \ref{eq-iou} where tp, fp, fn denote true positives, false positives, and false negatives respectively. As for multi-class segmentation mean class IoU, per-class IoU, mean category IoU and per-category IoU is used. Note that category IoU considers only category of classes instead of the specific classes when computing tp, fp and fn.

\begin{equation}
    precision= \frac{tp}{tp+fp}, 
    recall= \frac{tp}{tp+fn}
    \label{eq-prec}
\end{equation}

\begin{equation}
    F-measure= \frac{2*precision*recall}{precision+recall}
    \label{eq-fmes}
\end{equation}
\begin{equation}
    IoU = \frac{tp}{tp+fp+fn}
    \label{eq-iou}
\end{equation}

\begin{figure*}[ht!]
\centering
\begin{subfigure}{.15\textwidth}
    \centering
    \includegraphics[scale= 0.5]{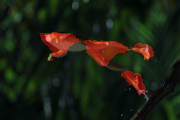}
\end{subfigure}%
\begin{subfigure}{.15\textwidth}
    \centering
    \includegraphics[scale= 0.5]{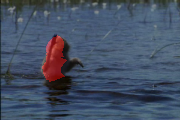}
\end{subfigure}%
\begin{subfigure}{.15\textwidth}
    \centering
    \includegraphics[scale= 0.5]{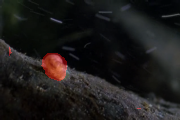}
\end{subfigure}%
\begin{subfigure}{.15\textwidth}
    \centering
    \includegraphics[scale= 0.5]{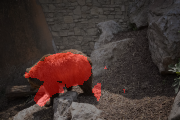}
\end{subfigure}%
\begin{subfigure}{.15\textwidth}
    \centering
    \includegraphics[scale= 0.5]{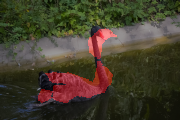}
\end{subfigure}%
\begin{subfigure}{.15\textwidth}
    \centering
    \includegraphics[scale= 0.5]{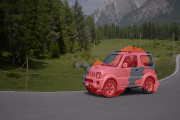}
\end{subfigure}

\begin{subfigure}{.15\textwidth}
    \centering
    \includegraphics[scale= 0.5]{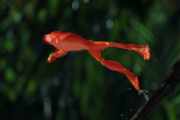}
\end{subfigure}%
\begin{subfigure}{.15\textwidth}
    \centering
    \includegraphics[scale= 0.5]{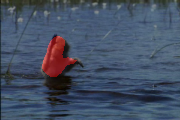}
\end{subfigure}%
\begin{subfigure}{.15\textwidth}
    \centering
    \includegraphics[scale= 0.5]{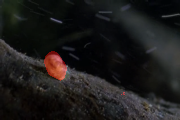}
\end{subfigure}%
\begin{subfigure}{.15\textwidth}
    \centering
    \includegraphics[scale= 0.5]{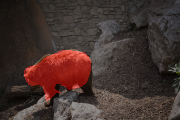}
\end{subfigure}%
\begin{subfigure}{.15\textwidth}
    \centering
    \includegraphics[scale= 0.5]{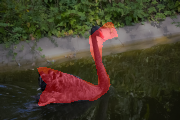}
\end{subfigure}%
\begin{subfigure}{.15\textwidth}
    \centering
    \includegraphics[scale= 0.5]{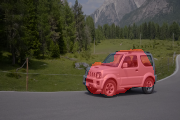}
\end{subfigure}
\caption{Qualitative results of experiments with SegtracV2 and Davis datasets, where network prediction are overlaid on the input. The top row is for FC-VGG and the bottom row is for RFC-VGG.}
\label{fig:qual}
\end{figure*}

\newcolumntype{L}[1]{>{\raggedright\arraybackslash\hspace{0pt}}p{#1}}
{\setlength{\extrarowheight}{10pt}
\begin{table*}[ht!]
\centering
\caption{Comparison of RFC-VGG with its baseline counterpart on DAVIS and SegTrack}
\label{table:segtrackdavis}
\begin{tabular}{|l|L{3cm}|c|c|c|c|}
\hline
\multicolumn{2}{|l|}{} & Precision & Recall & F-measure & IoU \\ \hline
\multirow{3}{*}{ SegTrack V2} &  FC-VGG & 0.7759 & 0.6810 & 0.7254 & 0.7646 \\ \cline{2-6} 
                   & RFC-VGG & \textbf{0.8325} & 0.7280 & \textbf{0.7767} & \textbf{0.8012}\\ \cline{2-6} 
                   & FC-VGG Extra Conv & 0.7519 & \textbf{0.7466} & 0.7493 & 0.7813 \\ \hline
\multirow{2}{*}{ DAVIS} & FC-VGG & 0.6834 & 0.5454 & 0.6066 & 0.6836 \\ \cline{2-6} 
                   & RFC-VGG & \textbf{0.7233} & \textbf{0.5586} & \textbf{0.6304} & \textbf{0.6984} \\ \hline
\end{tabular}
\end{table*}
In the first set pf experiments conducted, a fully convolutional VGG is used as a baseline denoted as FC-VGG and is compared against the recurrent version RFC-VGG. To avoid overfitting, first five layers of the network are initialized with the weights of a pre-trained networked and only lightly tuned. Table \ref{table:segtrackdavis} shows the results of the experiments on SegTrackV2 and DAVIS datasets. In these experiments, the data is split into half for training and the other half as keep out test set. RFC-VGG outperforms the FC-VGG architecture on both datasets with about 3\% and 5\% on DAVIS and SegTrack respectively. A comparison between using RFC-VGG versus using an extra convolutional layer with the same filter size (FC-VGG Extra Conv) is also presented. This result ensures that using the recurrent network to employ temporal data is the reason for the boost of performance not just merely adding extra convolutional filters. 

Figure \ref{fig:qual} shows the qualitative  analysis of RFC-VGG against FC-VGG. It shows that utilizing temporal information through the recurrent unit gives better segmentation for the object. This can be contributed to the implicit learning of the motion of segmented objects in the recurrent units. It also shows that using conv-GRU as the recurrent unit enables the extraction of temporal information from feature maps.
Note that the performance of the RFCN network depends on its baseline fully convolutional network. Thus, RFCN networks can be seen as a method to improve their baseline segmentation network by embedding them into a recurrent module that utilizes temporal data.

\begin{figure*}[ht!]
\centering
\begin{subfigure}{.25\textwidth}
    \centering
    \includegraphics[width=0.20\paperwidth]{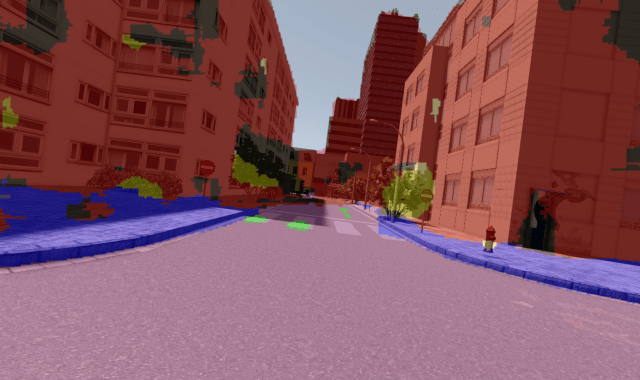}
\end{subfigure}%
\begin{subfigure}{.25\textwidth}
    \centering
    \includegraphics[width=0.20\paperwidth]{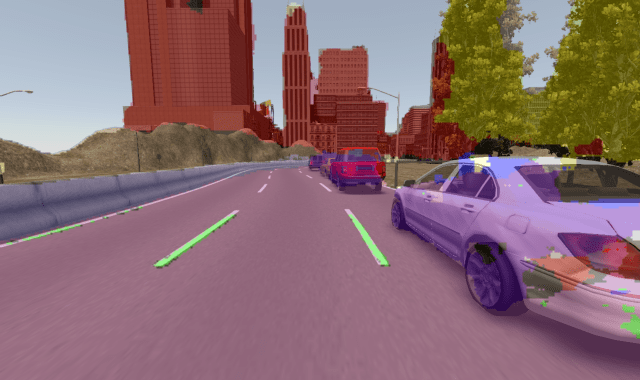}
\end{subfigure}%
\begin{subfigure}{.25\textwidth}
    \centering
    \includegraphics[width=0.20\paperwidth]{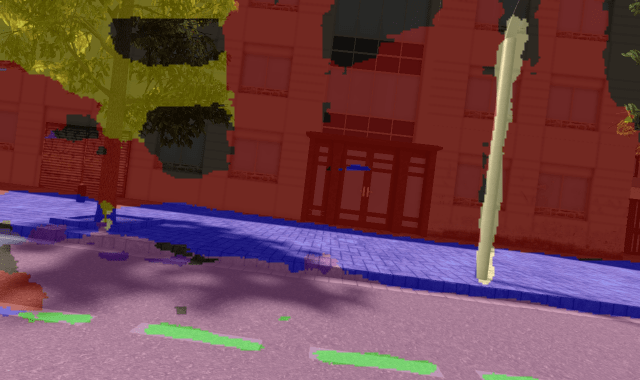}
\end{subfigure}%
\begin{subfigure}{.25\textwidth}
    \centering
    \includegraphics[width=0.20\paperwidth]{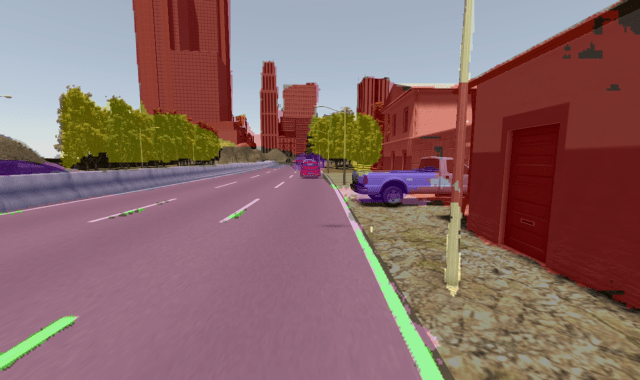}
\end{subfigure}%

\begin{subfigure}{.25\textwidth}
    \centering
    \includegraphics[width=0.20\paperwidth]{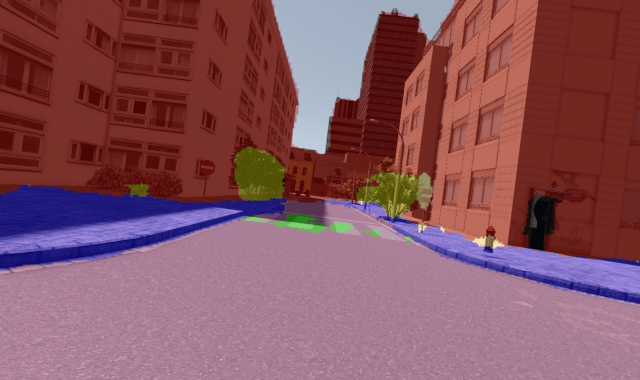}
\end{subfigure}%
\begin{subfigure}{.25\textwidth}
    \centering
    \includegraphics[width=0.20\paperwidth]{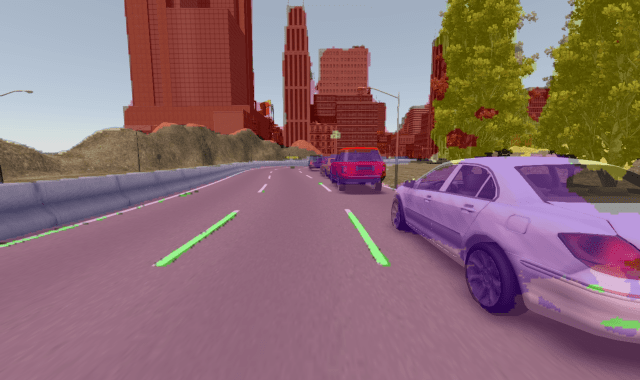}
\end{subfigure}%
\begin{subfigure}{.25\textwidth}
    \centering
    \includegraphics[width=0.20\paperwidth]{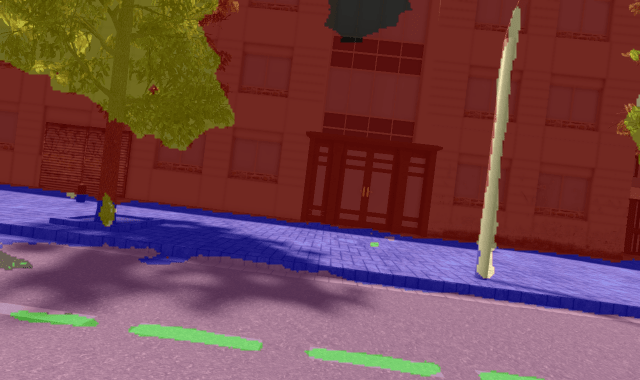}
\end{subfigure}%
\begin{subfigure}{.25\textwidth}
    \centering
    \includegraphics[width=0.20\paperwidth]{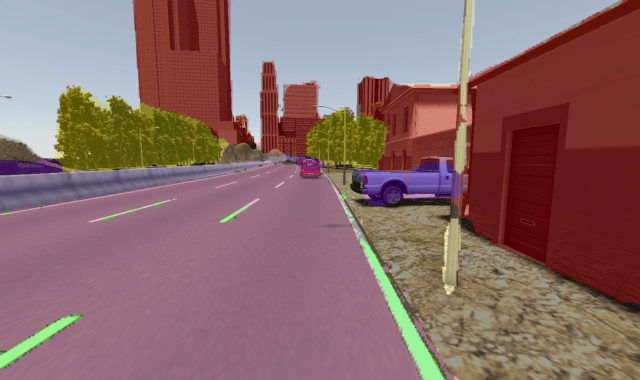}
\end{subfigure}%
\vspace{.2cm}

\begin{subfigure}{.25\textwidth}
    \centering
    \includegraphics[width=0.20\paperwidth]{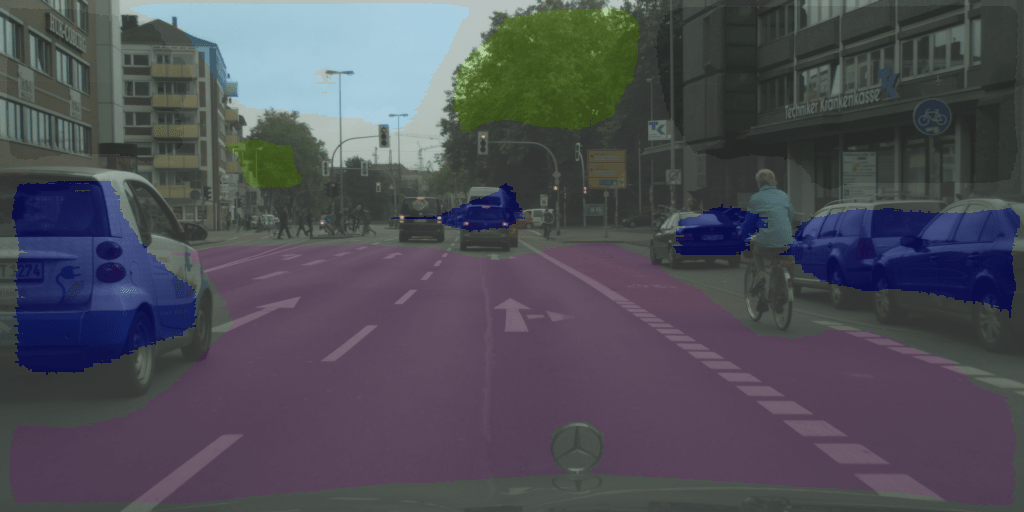}
\end{subfigure}%
\begin{subfigure}{.25\textwidth}
    \centering
    \includegraphics[width=0.20\paperwidth]{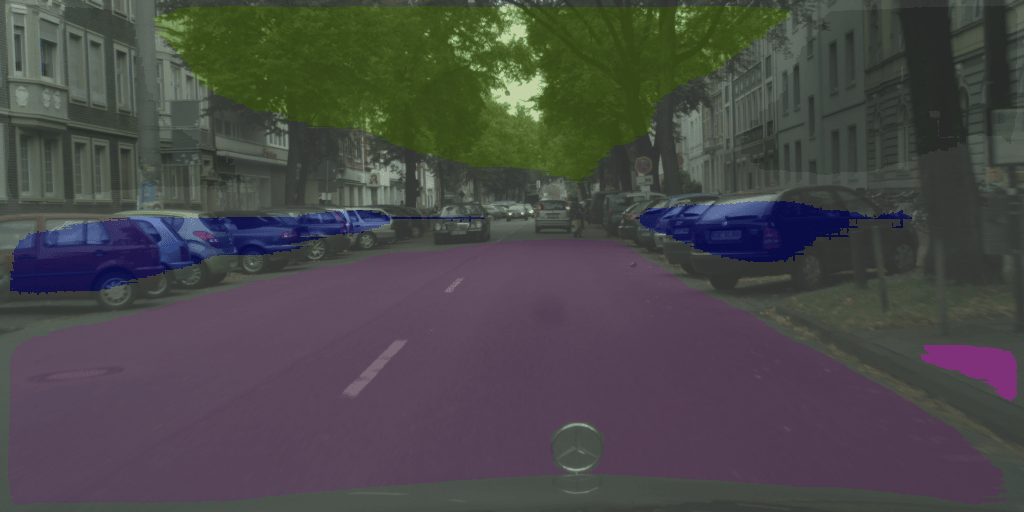}
\end{subfigure}%
\begin{subfigure}{.25\textwidth}
    \centering
    \includegraphics[width=0.20\paperwidth]{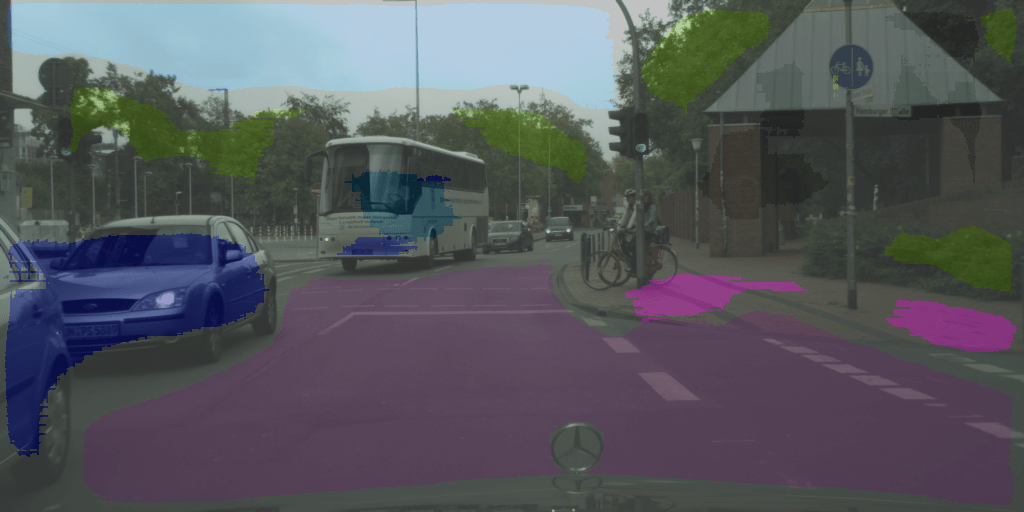}
\end{subfigure}%
\begin{subfigure}{.25\textwidth}
    \centering
    \includegraphics[width=0.20\paperwidth]{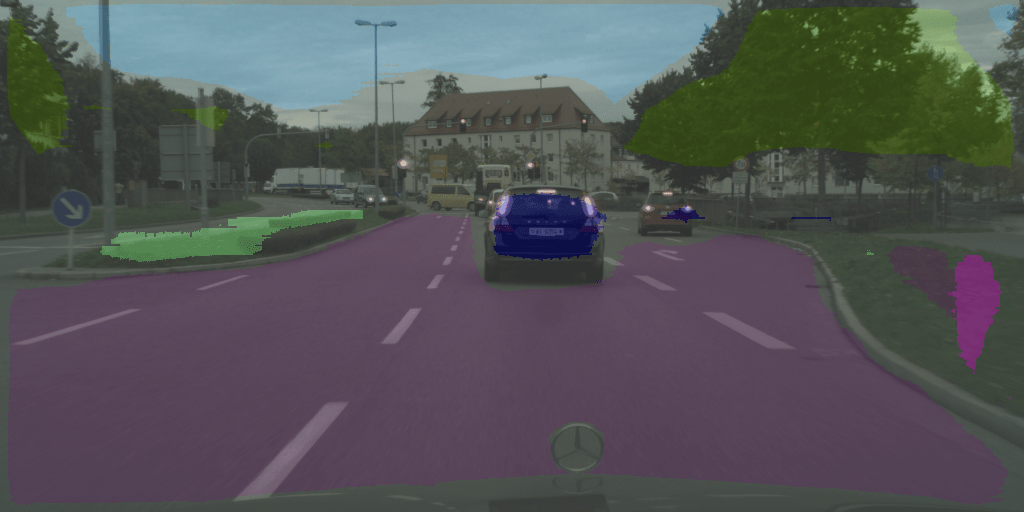}
\end{subfigure}%

\begin{subfigure}{.25\textwidth}
    \centering
    \includegraphics[width=0.20\paperwidth]{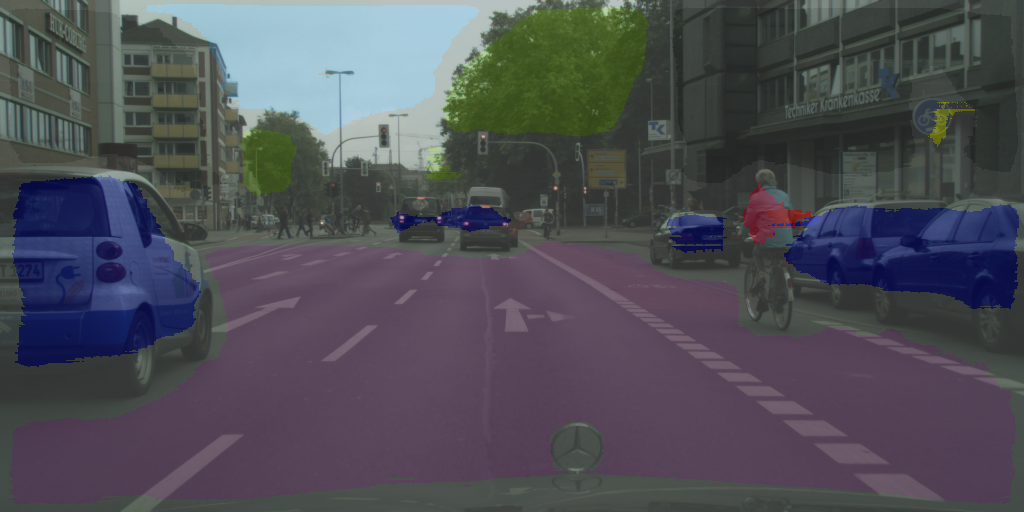}
\end{subfigure}%
\begin{subfigure}{.25\textwidth}
    \centering
    \includegraphics[width=0.20\paperwidth]{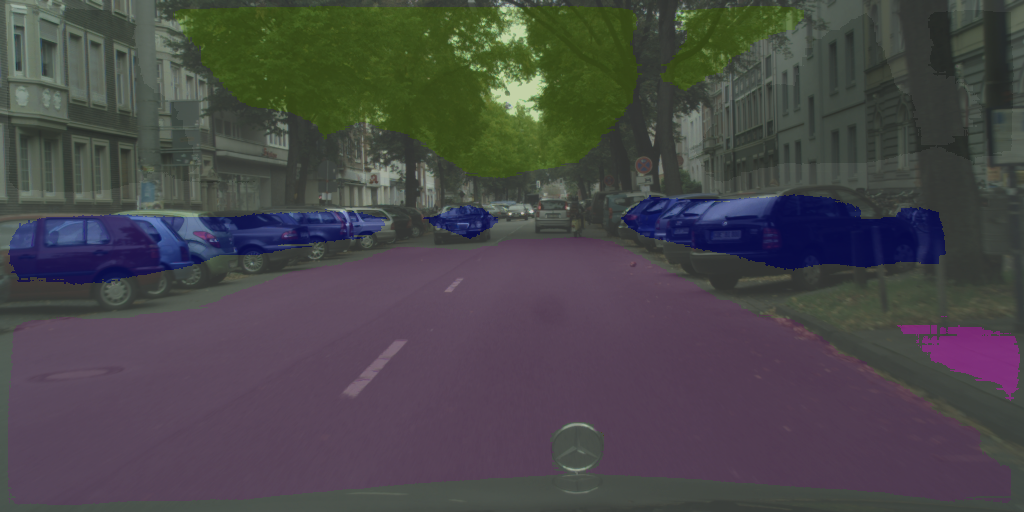}
\end{subfigure}%
\begin{subfigure}{.25\textwidth}
    \centering
    \includegraphics[width=0.20\paperwidth]{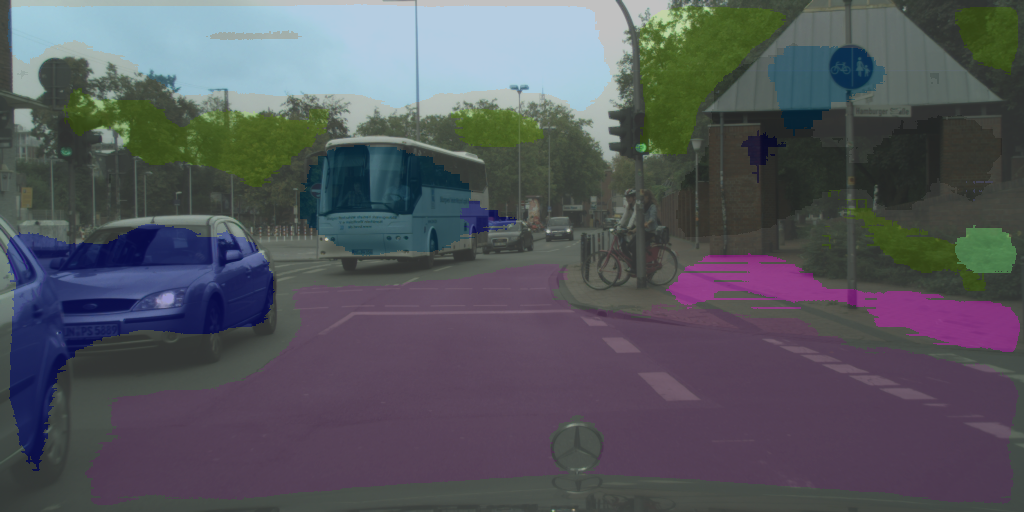}
\end{subfigure}%
\begin{subfigure}{.25\textwidth}
    \centering
    \includegraphics[width=0.20\paperwidth]{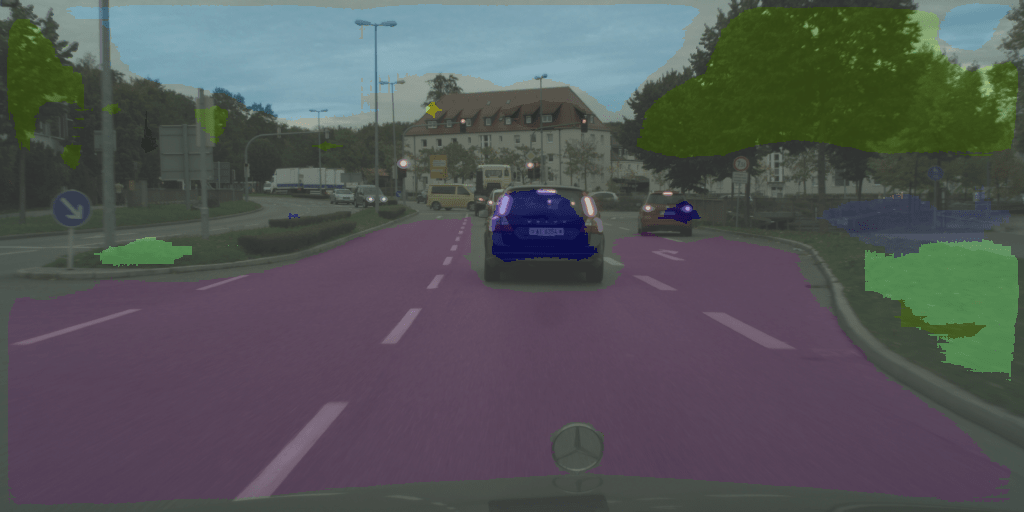}
\end{subfigure}%
\caption{Qualitative results of experiments with Synthia and cityscapes datasets, where network prediction are overlaid on the input. First row: Synthia with FC-VGG. Second row: Synthia with RFC-VGG. Third row: CityScapes with FCN-8s. Forth row: CityScapes with RFCN-8s:  }
\label{fig:qual_city}
\end{figure*}

\begin{table*}[ht!]
\centering
\caption{Semantic Segmentation Results on Synthia Highway Summer Sequence for RFC-VGG compared to FC-VGG}
\label{table:synthia}
\begin{tabular}{|c|c|c|c|c|c|c|c|c|c|c|}
\hline
\multirow{2}{*}{ } & \multirow{2}{*}{Mean Class IoU} & \multicolumn{9}{c|}{Per-Class IoU} \\ \cline{3-11} 
& & Car & Pedestrian & Sky  &  Building & Road & Sidewalk & Fence & Vegetation & Pole\\ \hline
FC-VGG & 0.755 & 0.504 &  0.275 & 0.946 &  0.958 & 0.840 & 0.957 & 0.762 & 0.883 & 0.718\\ \hline
RFC-VGG & \textbf{0.812} & \textbf{0.566} & \textbf{0.487} & \textbf{0.964} & \textbf{0.961} & \textbf{0.907} & \textbf{0.968} & \textbf{0.865} & \textbf{0.909} & \textbf{0.742}\\ \hline
\end{tabular}
\end{table*}

\begin{table*}[ht!]
\centering
\caption{Semantic Segmentation Results on CityScapes for RFCN-8s compared to FCN-8s}
\label{table:cityscapes}
\begin{tabular}{|c|c|c|c|c|c|c|}
\hline
\multirow{2}{*}{ } & \multirow{2}{*}{Category IoU} & \multicolumn{5}{c|}{Per-category IoU} \\ \cline{3-7} 
&  &  Flat & Nature   & Sky & Construction & Vehicle \\ \hline
\multicolumn{1}{|c|}{FCN-8s} & 0.53 &  0.917 & 0.710 &  0.792 & 0.683 & 0.585 \\ \hline
RFCN-8s  & \textbf{0.565}  & \textbf{0.928} &  \textbf{0.739} & \textbf{0.814} & \textbf{0.719} & \textbf{0.652} \\ \hline
\end{tabular}
\end{table*}

The same architecture was used for semantic segmentation on synthia dataset after modifying it to support the thirteen classes. A comparison between FC-VGG and RFC-VGG is presented in terms of mean class IoU and per-class IoU for some of the classes. Table\ref{table:synthia} presents the results on synthia dataset. RFC-VGG has 5.7\% over FC-VGG in terms of mean class IoU. It also shows the per-class IoU generally improves in the case of RFC-VGG. Interestingly, the highest improvement is with the car and pedestrian class that benefits the most from a learned motion pattern compared to sky or buildings that are mostly static. Figure\ref{fig:qual_city} first row shows the qualitative analysis on Synthia. The second image shows the car's enhanced segmentation with RFC-VGG.

Finally, experimental results on cityscapes dataset using FCN-8s and its recurrent version RFCN-8s is shown in Table\ref{table:cityscapes}. It uses mean category IoU and per-category IoU for the evaluation. It clearly demonstrates that RFCN-8s outperforms FCN-8s with 3.5\% on mean category IoU. RFCN-8s generally improves on the per-category IoU, with the highest improvement in vehicle category. Hence, again the highest improvement is in the category that is affected the most by temporal data. Figure \ref{fig:qual_city} bottom row shows the qualitative evaluation on cityscapes data to compare FCN-8s versus RFCN-8s. The third image clearly shows that the moving bus is better segmented with the recurrent version. Note that the experiments were conducted on images with less resolution than the original data and with a reduced version of FCN-8s due to memory constraints. Therefore, finer categories such as human and object are poorly segmented. However, using original resolution will fix this problem and its recurrent version should have better results as well.

\subsection{Additional Experiments}
In this section, experiments using conventional recurrent layers for segmentation is presented. These experiments provide further analysis on different recurrent units and their effects on the RFCN. A comparison between end-to-end training versus the decoupled one is also presented. The moving MNIST and change detection datasets are used for this part. Images of MNIST dataset are relatively small (28$\times$28) which allows us to test our RFC-Lenet network \ref{table:networks}. A fully convolutional Lenet is compared against RFC-Lenet. Table \ref{table_mnist} shows the results that were obtained. The results of RFC-Lenet with GRU is better than FC-Lenet with 2\% improvement. Note that GRU gave better results than LSTM in as well.
\\
\begin{table}[!htb]
  \centering
  \caption{Precision, Recall, and F-measure on FC-Lenet, LSTM, GRU, and RFC-Lenet tested on the moving MNIST dataset}
  \begin{tabular}{| L{2cm} | C{1.5cm} | C{1.5cm} | C{1.5cm} |}
    \hline
     & Precision & Recall & F-measure\\ \hline
    FC-Lenet & 0.868 & \textbf{0.922} & 0.894\\ \hline
    LSTM & 0.941 & 0.786 & 0.856 \\ \hline
    GRU & 0.955 & 0.877 & 0.914 \\ \hline
    RFC-Lenet & \textbf{0.96} & 0.877 & \textbf{0.916}\\
    \hline
  \end{tabular}
 
 \label{table_mnist}
\end{table}
\\

We used real data from motion detection benchmark for the second set of experiments. The training and test splits are 70\% and 30\% from each sequence throughout these experiments. Baseline FC-12s is compared against its recurrent version, RFC-12s. 
It is also compared against the decoupled training of the FC-12s and the recurrent unit. Where GRU is trained on the heat map output from FC-12s. Table \ref{table_alldata} shows the results of these experiments, where the RFC-12s network had a 1.4\% improvement over FC-12s. We observe less relative improvement compared to using Conv-GRU because in regular GRU spatial connectivities are ignored. However, incorporating the temporal data still helped the segmentation accuracy.

\begin{table}[!htb]
  \centering
   \caption{Precision, Recall, and F-measure on motion detection dataset. FCN-12s is the baseline FCN network. RFC-12s is its counterpart recurrent network. RFC-12s is trained in two ways. Decoupled (D) where first, the GRU layer alone is trained with the rest of the network fixed and then the whole network finely tuned together. End-to-end (EE) where the whole network is trained at once.}
  \begin{tabular}{| L{2.5cm} | C{1.4cm} | C{1cm} | C{1.5cm} |}
   \hline
     & Precision & Recall & F-measure\\ \hline
    FC-12s & 0.827 & 0.585 & 0.685 \\ \hline
    RFC-12s (D) & \textbf{0.835} & 0.587 & 0.69\\ \hline
    RFC-12s (EE) & 0.797 & \textbf{0.623} & \textbf{0.7}\\
    \hline
  \end{tabular}
   \label{table_alldata}
\end{table}

\section{Conclusion and Future Work} \label{conclusion}
We presented a novel method that exploits implicit temporal information in videos to improve segmentation. This approach utilizes convolutional gated recurrent network which allows it to use preceding frames in segmenting the current frame. We performed extensive experiments on six datasets with different segmentation objective. We showed that embedding FCN networks as a recurrent module, consistently improved the results through out different datasets. Specifically, a 5\% improvement in Segtrack and 3\% improvement in Davis in F-measure over a plain fully convolutional network; a 5.7\% improvement on Synthia in mean IoU, and 3.5\% improvement on CityScapes in mean category IoU, over the plain fully convolutional network. Our suggested architecture can be applied on any FCN like single frame segmentation and then be used to process videos in an online fashion with an improved performance.\\

For future work, we would like to enhance the results of the semantic segmentation and apply our recurrent method to more single-image segmentation networks, for a more complete comparison with the state of the art. Another direction is to explore the potential of incorporating shape constraints from the depth data within the network. Thus combining motion and shape cues for better video segmentation.

{\small
\bibliographystyle{ieee}
\bibliography{egbib}

\begin{thebibliography}{10}\itemsep=-1pt

\bibitem{badrinarayanan2010label}
V.~Badrinarayanan, F.~Galasso, and R.~Cipolla.
\newblock Label propagation in video sequences.
\newblock In {\em Computer Vision and Pattern Recognition (CVPR), 2010 IEEE
  Conference on}, pages 3265--3272. IEEE, 2010.

\bibitem{ballas2015delving}
N.~Ballas, L.~Yao, C.~Pal, and A.~Courville.
\newblock Delving deeper into convolutional networks for learning video
  representations.
\newblock {\em arXiv preprint arXiv:1511.06432}, 2015.

\bibitem{Bastien-Theano-2012}
F.~Bastien, P.~Lamblin, R.~Pascanu, J.~Bergstra, I.~J. Goodfellow, A.~Bergeron,
  N.~Bouchard, and Y.~Bengio.
\newblock Theano: new features and speed improvements.
\newblock Deep Learning and Unsupervised Feature Learning NIPS 2012 Workshop,
  2012.

\bibitem{bengio1994learning}
Y.~Bengio, P.~Simard, and P.~Frasconi.
\newblock Learning long-term dependencies with gradient descent is difficult.
\newblock {\em Neural Networks, IEEE Transactions on}, 5(2):157--166, 1994.

\bibitem{cho2014properties}
K.~Cho, B.~van Merri{\"e}nboer, D.~Bahdanau, and Y.~Bengio.
\newblock On the properties of neural machine translation: Encoder-decoder
  approaches.
\newblock {\em arXiv preprint arXiv:1409.1259}, 2014.

\bibitem{chung2014empirical}
J.~Chung, C.~Gulcehre, K.~Cho, and Y.~Bengio.
\newblock Empirical evaluation of gated recurrent neural networks on sequence
  modeling.
\newblock {\em arXiv preprint arXiv:1412.3555}, 2014.

\bibitem{cordts2016cityscapes}
M.~Cordts, M.~Omran, S.~Ramos, T.~Rehfeld, M.~Enzweiler, R.~Benenson,
  U.~Franke, S.~Roth, and B.~Schiele.
\newblock The cityscapes dataset for semantic urban scene understanding.
\newblock {\em arXiv preprint arXiv:1604.01685}, 2016.

\bibitem{donahue2015long}
J.~Donahue, L.~Anne~Hendricks, S.~Guadarrama, M.~Rohrbach, S.~Venugopalan,
  K.~Saenko, and T.~Darrell.
\newblock Long-term recurrent convolutional networks for visual recognition and
  description.
\newblock In {\em Proceedings of the IEEE Conference on Computer Vision and
  Pattern Recognition}, pages 2625--2634, 2015.

\bibitem{goyette2012changedetection}
N.~Goyette, P.-M. Jodoin, F.~Porikli, J.~Konrad, and P.~Ishwar.
\newblock Changedetection. net: A new change detection benchmark dataset.
\newblock In {\em Computer Vision and Pattern Recognition Workshops (CVPRW),
  2012 IEEE Computer Society Conference on}, pages 1--8. IEEE, 2012.

\bibitem{hochreiter1997long}
S.~Hochreiter and J.~Schmidhuber.
\newblock Long short-term memory.
\newblock {\em Neural computation}, 9(8):1735--1780, 1997.

\bibitem{johnson2015densecap}
J.~Johnson, A.~Karpathy, and L.~Fei-Fei.
\newblock Densecap: Fully convolutional localization networks for dense
  captioning.
\newblock {\em arXiv preprint arXiv:1511.07571}, 2015.

\bibitem{li2013video}
F.~Li, T.~Kim, A.~Humayun, D.~Tsai, and J.~M. Rehg.
\newblock Video segmentation by tracking many figure-ground segments.
\newblock In {\em Proceedings of the IEEE International Conference on Computer
  Vision}, pages 2192--2199, 2013.

\bibitem{lin2015efficient}
G.~Lin, C.~Shen, I.~Reid, et~al.
\newblock Efficient piecewise training of deep structured models for semantic
  segmentation.
\newblock {\em arXiv preprint arXiv:1504.01013}, 2015.

\bibitem{long2015fully}
J.~Long, E.~Shelhamer, and T.~Darrell.
\newblock Fully convolutional networks for semantic segmentation.
\newblock In {\em Proceedings of the IEEE Conference on Computer Vision and
  Pattern Recognition}, pages 3431--3440, 2015.

\bibitem{miksik2015semantic}
O.~Miksik, V.~Vineet, M.~Lidegaard, R.~Prasaath, M.~Nie{\ss}ner, S.~Golodetz,
  S.~L. Hicks, P.~P{\'e}rez, S.~Izadi, and P.~H. Torr.
\newblock The semantic paintbrush: Interactive 3d mapping and recognition in
  large outdoor spaces.
\newblock In {\em Proceedings of the 33rd Annual ACM Conference on Human
  Factors in Computing Systems}, pages 3317--3326. ACM, 2015.

\bibitem{noh2015learning}
H.~Noh, S.~Hong, and B.~Han.
\newblock Learning deconvolution network for semantic segmentation.
\newblock In {\em Proceedings of the IEEE International Conference on Computer
  Vision}, pages 1520--1528, 2015.

\bibitem{pavel2015recurrent}
M.~S. Pavel, H.~Schulz, and S.~Behnke.
\newblock Recurrent convolutional neural networks for object-class segmentation
  of rgb-d video.
\newblock In {\em Neural Networks (IJCNN), 2015 International Joint Conference
  on}, pages 1--8. IEEE, 2015.

\bibitem{perazzibenchmark}
F.~Perazzi, J.~P.-T.~B. McWilliams, L.~Van~Gool, M.~Gross, and
  A.~Sorkine-Hornung.
\newblock A benchmark dataset and evaluation methodology for video object
  segmentation.

\bibitem{perazzi2015fully}
F.~Perazzi, O.~Wang, M.~Gross, and A.~Sorkine-Hornung.
\newblock Fully connected object proposals for video segmentation.
\newblock In {\em Proceedings of the IEEE International Conference on Computer
  Vision}, pages 3227--3234, 2015.

\bibitem{ramakanth2014seamseg}
S.~A. Ramakanth and R.~V. Babu.
\newblock Seamseg: Video object segmentation using patch seams.
\newblock In {\em 2014 IEEE Conference on Computer Vision and Pattern
  Recognition}, pages 376--383. IEEE, 2014.

\bibitem{ros2016synthia}
G.~Ros, L.~Sellart, J.~Materzynska, D.~Vazquez, and A.~M. Lopez.
\newblock The synthia dataset: A large collection of synthetic images for
  semantic segmentation of urban scenes.
\newblock In {\em Proceedings of the IEEE Conference on Computer Vision and
  Pattern Recognition}, pages 3234--3243, 2016.

\bibitem{simonyan2014very}
K.~Simonyan and A.~Zisserman.
\newblock Very deep convolutional networks for large-scale image recognition.
\newblock {\em arXiv preprint arXiv:1409.1556}, 2014.

\bibitem{vineet2015icra}
V.~Vineet, O.~Miksik, M.~Lidegaard, M.~Nie{\ss}ner, S.~Golodetz, V.~A.
  Prisacariu, O.~K\"ahler, D.~W. Murray, S.~Izadi, P.~Perez, and P.~H.~S. Torr.
\newblock Incremental dense semantic stereo fusion for large-scale semantic
  scene reconstruction.
\newblock In {\em IEEE International Conference on Robotics and Automation
  (ICRA)}, 2015.

\bibitem{vinyals2012revisiting}
O.~Vinyals, S.~V. Ravuri, and D.~Povey.
\newblock Revisiting recurrent neural networks for robust asr.
\newblock In {\em Acoustics, Speech and Signal Processing (ICASSP), 2012 IEEE
  International Conference on}, pages 4085--4088. IEEE, 2012.

\bibitem{visin2015reseg}
F.~Visin, K.~Kastner, A.~Courville, Y.~Bengio, M.~Matteucci, and K.~Cho.
\newblock Reseg: A recurrent neural network for object segmentation.
\newblock {\em arXiv preprint arXiv:1511.07053}, 2015.

\bibitem{wolf2016enhancing}
D.~Wolf, J.~Prankl, and M.~Vincze.
\newblock Enhancing semantic segmentation for robotics: The power of 3-d
  entangled forests.
\newblock {\em IEEE Robotics and Automation Letters}, 1(1):49--56, 2016.

\bibitem{zeiler2012adadelta}
M.~D. Zeiler.
\newblock Adadelta: an adaptive learning rate method.
\newblock {\em arXiv preprint arXiv:1212.5701}, 2012.

\bibitem{zhang2013understanding}
H.~Zhang, A.~Geiger, and R.~Urtasun.
\newblock Understanding high-level semantics by modeling traffic patterns.
\newblock In {\em Proceedings of the IEEE International Conference on Computer
  Vision}, pages 3056--3063, 2013.

\bibitem{zheng2015conditional}
S.~Zheng, S.~Jayasumana, B.~Romera-Paredes, V.~Vineet, Z.~Su, D.~Du, C.~Huang,
  and P.~H. Torr.
\newblock Conditional random fields as recurrent neural networks.
\newblock In {\em Proceedings of the IEEE International Conference on Computer
  Vision}, pages 1529--1537, 2015.

\end{thebibliography}
}

\end{document}